# Breaking the Homogeneity Assumption: Specialized Multi-Generator Adversarial Learning for Rare Failure Detection in Predictive Maintenance

**Alexis Lazanas**
*Department of Mechanical Engineering and Aeronautics, University of Patras, Patras, Greece*
alexlas@upatras.gr

**Georgios Kampouropoulos**
*Department of Mechanical Engineering and Aeronautics, University of Patras, Patras, Greece*
up1067279@ac.upatras.gr

## ABSTRACT
Supervised learning models in the predictive maintenance field are regularly trained on industrial datasets which are highly imbalanced: machine failures occur rarely but have a disproportionate effect on operations. In addition to the clear class disparity, the data of failures are typically non-homogeneous, with the different modes of failure being based on different physical processes and having a multimodal distribution among the minorities and the classes. Traditional imbalance-management methods e.g. undersampling, SMOTE-based interpolation or cost-sensitive learning, typically assume that the minority population is a homogeneous, homogenous group. This means that their effectiveness is severely limited in multifaceted conditions that are experienced in industrial practice. This paper determines the possibility of a failure-type-conscious generative augmentation program to improve the identification of infrequent failures in predictive maintenance systems. An experimental design that is leakage-safe is used to compare five imbalance-handling methods: cost-sensitive learning, random undersampling, SMOTE oversampling, single-generator GAN augmentation, and a specialized multi-generator GAN architecture that has independent generators that are asked to learn individual failure subtypes. Precision/Recall-oriented measures are used to quantify model performance, the main evaluation measure is the PR-AUC. Experiments carried out on the AI4I 2020 predictive maintenance dataset indicate that the suggested multi-generator GAN framework produces more realistic samples of minorities, thus producing better PR-AUC and recall scores in comparison to traditional resampling methods and individual-generator GAN augmentation. Though this method comes at the cost of higher computational costs, the findings provide strong evidence that generator specialization is a more efficient method to cope with the heterogeneous distributions of failure, that are inherent to the imbalanced predictive maintenance cases.

## 1. INTRODUCTION
In Industry 4.0 environments, PdM has emerged as a core element of industrial analytics, in an effort to minimize unplanned downtimes, increase of assets' lifespan, and facilitate data-informed maintenance decisions. Modern PdM systems are based on supervised learning models that are trained on historical sensor and operational data to separate normal operating conditions and equipment failures [1], [36]. Although sensing and analytics have made progress, the usefulness of these systems is frequently limited by statistical characteristics of industrial data instead of models' capacity per se.

One issue with PdM that has been persistent is severe class imbalance, with failure events being infrequent but disproportionately important in terms of operational value. This leads to learning imbalance that favors majority group behavior and makes evaluation difficult because conventional measures like accuracy in general can mask poor performance on the minority failure group [6], [12]. As a result, class-sensitive evaluation measures, such as precision, recall, and area under the precision-recall curve (PR-AUC), are becoming more common in PdM research papers, which more directly reflect the failure detection performance in skewed class distributions.

Beyond imbalance severity, industrial failure data is often heterogeneous in nature. Failures are usually caused by different physical mechanisms, e.g., mechanical wear, thermal stress, electrical faults, etc. Minority class distributions are not homogeneous but rather multimodal. Conventional data-level imbalance handling techniques such as random undersampling and synthetic oversampling generally treat the minority class as a single united population. Random undersampling tends to be more failure-sensitive, but at the cost of loss of informative samples of the majority-class, whereas any SMOTE-based approach may add noise or blur the decision boundaries due to the use of linear interpolation in high-dimensional feature spaces [13], [12]. These limitations are more pronounced by the presence of substantial difference between the underlying data distributions of failure modes.

Recent studies consider generative adversarial networks (GANs) in order to augment imbalanced data by learning more non-linear representations of minority-class data. GAN-based oversampling proved to be useful in PdM, where synthetic failure samples are generated, which more closely reproduce complex data manifolds compared to classical interpolation-based approaches [14], [45]. Nevertheless, much of the available methods use a single generator to import all failure data in a single model, implicitly assuming that distributions are coherent across failure types [49]. This assumption should not be assumed in a heterogeneous industrial environment where generative models might not be able to capture the characteristics of failures.

Based on the above, this paper examines the possibility of using failure-type-conscious generative oversampling to enhance classification in an imbalanced PdM task. The paper assesses a multi-generator GAN model where individual generators are trained to capture different subtypes of failures where failure-specific structure should be maintained during data augmentation. The proposed approach is evaluated with the cost-sensitive learning, random undersampling, SMOTE oversampling, and single generator GAN augmentation with leakage safe experimental protocol. Performance is measured using precision-recall metrics that are suitable for rare event industrial classification problems. This work aims to explore the application of structured oversampling in line with the heterogeneity of failures in order to offer empirical evidence on how specialized generative augmentation can be applied to ensure robust failure detection in machine-learning pipelines in the industrial sector.

The rest of this paper is structured as follows: Section 2 provides an overview of recent literature on PdM under class imbalance, focusing on data-level resampling approaches and generative methods, and the issues in methods that assume homogenous failure distributions. Section 3 explains the methodological research design, data handling, test procedure, and performance measures that are employed in order to make leakage-safe and fair comparisons. Section 4 presents the experimental results and comparative performance analysis for all the sampling strategies. Finally, Section 5 concludes the paper and provides directions for future research.

## 2. RELATED WORK

Recent research in the area of PdM has repeatedly emphasized that the efficiency of data-driven failure detection systems is often limited by the statistical nature of industrial data sets rather than by the expressiveness of the model used. Detailed surveys show that extreme class distribution, lack of failure data, and complicated operating regimes are the key problems in all fields of industrial activity, even with the development of sensing, connectivity, and learning algorithms [1], [36], [35]. As a result, a large amount of work has been done on imbalance-aware learning strategies for PdM applications.

Algorithmic approaches like cost-sensitive learning and class-weighted losses, are often used to address the imbalance effects without altering the training data. While these methods are fast and provide accurate data distribution retention, recent comparative studies suggest that their effectiveness decreases when failure samples are sparse or poorly separated in feature space, which is a typical scenario in the industry [6], [43]. Consequently, the data-level resampling techniques still have a prominent role in PdM pipelines.

Random undersampling is considered a common baseline due to its simplicity, however, several works report that the random removal of samples from the majority classes can degrade the performance of classifiers by removing informative operating conditions, especially in multimodal systems [12]. Synthetic oversampling methods, and in particular SMOTE and its derivatives, have thus been widely tested and used in PdM. Though these techniques can be used to enhance minority-class recall, recent large-scale experiments demonstrate that interpolation-based oversampling can create synthetic samples in areas of class overlap or to decision boundaries and cannot be used with high-dimensional or heterogeneous failure conditions [13], [12], [33], [50].

To overcome these limitations, generative models have become more and more popular as an alternative to minority class augmentation. GANs, specifically, have been studied to generate failure data in industrial fault detection and condition monitoring tasks and a number of papers have demonstrated their increased effectiveness over traditional oversampling methods [14], [26]. In addition to PdM, wider studies of imbalanced learning verify that GAN-based augmentation can more effectively represent complex, non-linear minority distributions than linear interpolation methods, particularly in situations where the boundary of the classes is fragmented or very jagged [17], [23].

Despite their promise, current GAN-based oversampling methods in PdM mostly utilize single-generator formulations implicitly based on the assumption that all failure cases can be treated as a single distribution. This assumption has been challenged in recent research showing that industrial failures often stem from different physical mechanisms and have heterogeneous statistical characteristics [26], [35]. Conditional GANs have been suggested as a partial solution to the problem by taking class labels into account during the generation phase, which allows more controlled synthesis of minority samples [19]. Nevertheless, conditioning is not always a guarantee of failure substructure modeling, especially when the individual failure modes are highly imbalanced themselves.

Recent reviews bring into focus the imbalance handling methods that avoid global distributional assumptions but maintain minority-class substructure [12], [35]. However, there is little systematic assessment of generative oversampling methods that explicitly consider the heterogeneity of failure in the PdM literature. In particular, there are few studies investigating if generative models can be separated based on failure subtype to improve learning for minority classes with leakage-safe experimental protocols and evaluation schemes that are centered on precision-recall performance [2], [9]. This gap serves as the motivation for this study, where the proposed failure-type-aware generative oversampling is a structured alternative to the homogeneous resampling approaches in the case of imbalanced PdM classification.

## 3. IMPLEMENTATION AND EXPERIMENTAL FRAMEWORK

This Section details the methodological and experimental framework followed in the research. The primary focus is the construction of a robust, leakage-safe data processing pipeline and the subsequent application of various sampling strategies to address class imbalance. We begin by describing the preprocessing and exploratory data analysis performed on the **AI4I 2020** dataset, which forms the foundation for all subsequent modeling scenarios. Each step is justified by connecting the practical implementation choices to the established best practices for PdM and imbalanced classification discussed in Chapter 2.

### 3.1 Data Preprocessing and Exploratory Data Analysis Normal or Body Text

The initial phase of the project involved a thorough preprocessing of the AI4I-2020 dataset to prepare it for classification [3]. This process included data auditing, feature engineering, encoding, and scaling, all orchestrated within a pipeline designed to prevent information leakage. Exploratory Data Analysis (EDA) was conducted in parallel to inform and validate these preprocessing decisions.

**Problem Setup and Notation:** The task is framed as a supervised binary classification problem, where the objective is to predict machine failure (*y=1*) versus normal operation (*y=0*) using sensor and operational data from a simulated manufacturing process [34]. The dataset exhibits a significant class imbalance, with a failure rate of approximately 3.4%, a common characteristic of real-world PdM applications [33], [26]. The initial dataset comprises 10,000 instances and 14 features. The unique identifiers, *UDI* and *Product_ID*, were immediately removed as they hold no predictive value and could lead to overfitting if treated as features. The final feature set consists of six numerical sensor readings and one categorical feature, **‘*Type*’**, which represents the quality of the machine.

**Table 1. Descriptive statistics before preprocessing (AI4I.**

| Statistic | AirTemp [K] | Process temp [K] | Rotational speed [rpm] | Torque [Nm] | Tool wear [min] | Target |
|---|---|---|---|---|---|---|
| **count** | 10000.00 | 10000.0 | 10000.0 | 10000.0 | 10000.0 | 10000.0 |
| **mean** | 300.00 | 310.00 | 1538.77 | 39.98 | 107.95 | 0.033 |
| **std** | 2.00 | 1.4837 | 179.284 | 9.9689 | 63.6541 | 0.1809 |
| **min** | 295.30 | 305.70 | 1168.00 | 3.8000 | 0.0000 | 0.0000 |
| **25%** | 298.30 | 308.80 | 1423.00 | 33.200 | 53.000 | 0.0000 |
| **50%** | 300.10 | 310.10 | 1530.00 | 40.1000 | 108.00 | 0.0000 |
| **75%** | 301.50 | 311.10 | 1612.00 | 46.800 | 162.000 | 0.0000 |
| **max** | 304.50 | 313.80 | 2886.00 | 76.6000 | 253.000 | 1.000 |

**Data Auditing and Missingness:** A preliminary audit of the dataset was conducted to check for data quality issues such as duplicates, constant features, or physically impossible values. The AI4I 2020 dataset was found to be exceptionally clean, with no duplicate rows or constant features. Most importantly, the dataset contained **no missing values**. Consequently, imputation strategies such as median, k-nearest neighbors, or iterative imputation, were not necessary for this project [22]. Had missingness been present, a robust policy such as median imputation for its resilience to outliers, or a more sophisticated method like KNN imputation, would have been implemented strictly within each cross-validation fold to prevent data leakage [46], [33].

**Feature Types and Encoding:** The dataset contains a single categorical feature, ‘***Type***’, which has three levels: ‘***L***’ (*Low*), ‘***M***’ (*Medium*), and ‘***H***’ (*High*) quality. These levels possess an inherent ordinal relationship (*L* < *M* < *H*). To preserve this meaningful order, an *OrdinalEncoder* was chosen over one-hot encoding. While one-hot encoding is a strong baseline for nominal data, it would have discarded this ordinal information and increased the dimensionality of the feature space [47]. By mapping ‘***L***’, ‘***M***’ and ‘***H***’ to integers (e.g., 0, 1, 2), we provide the model with a potentially useful signal regarding product quality’s impact on machine performance. This encoding step was integrated into a *ColumnTransformer* pipeline to ensure it was fitted consistently and without leakage within each cross-validation fold [46].

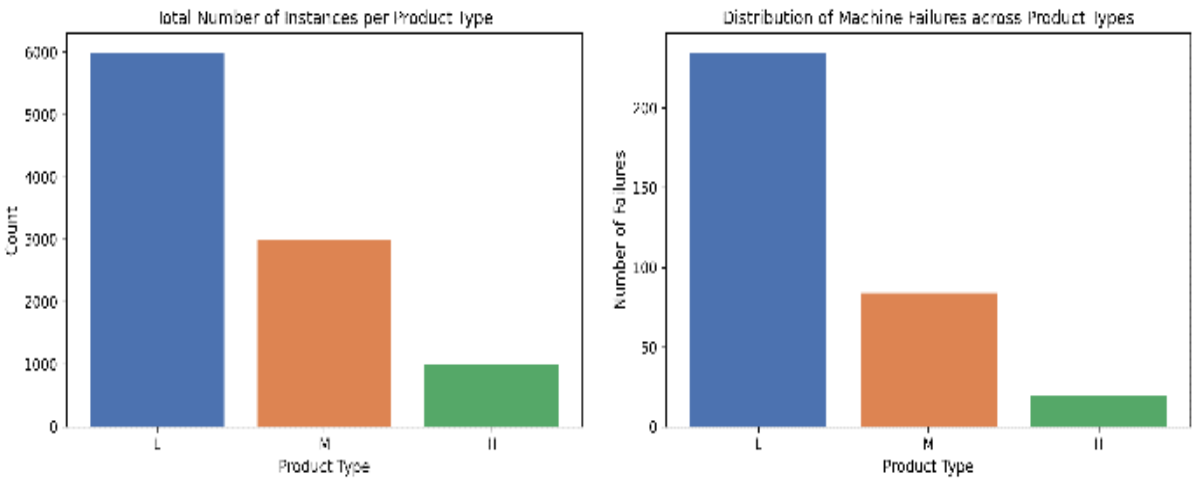

**Fig 1: Product type analysis. Left: total number of instances per product type (L, M, H). Right: number of machine failures across product types, showing failures concentrated in type L relative to M and H**

**Scaling and Distributional Transformation:** All numerical features were scaled using *StandardScaler*, which standardizes features by removing the mean and scaling to unit variance. This step is critical for distance-based algorithms like Support Vector Machines, whose performance can be severely degraded if features with large magnitudes are allowed to dominate the kernel calculations [37]. Although tree-based models like Random Forest are largely invariant to monotonic feature transformations, applying a consistent scaling across the entire feature set simplifies the experimental pipeline and ensures comparability between the two classifier families. While distributional transforms like the *Yeo-Johnson* [44] or *Box-Cox* [10] methods were considered to handle potential skewness, an initial review of the feature distributions revealed that they were not sufficiently skewed to warrant the additional complexity. Therefore, a simple standardization was deemed a robust and sufficient baseline for this thesis. The temperature grades were converted from Kelvin to Celsius for easier user validation.

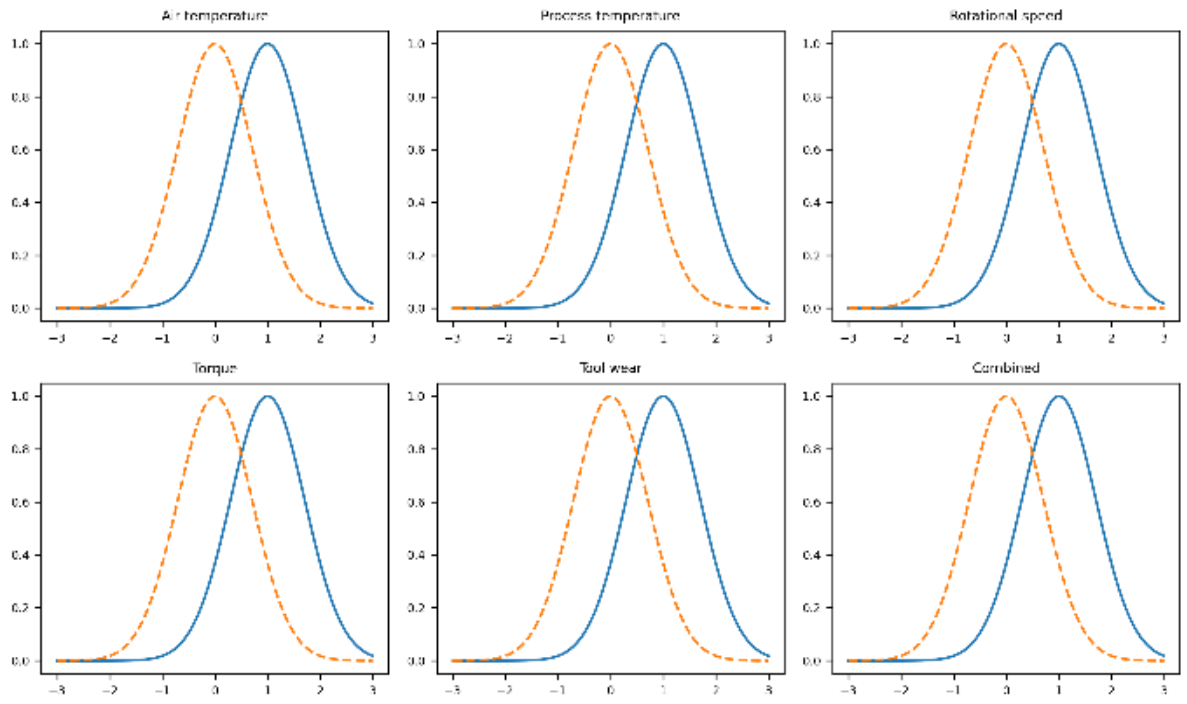

**Fig 2: Distribution of numerical features before (blue line) and after standardization (orange-dotted line). The scaler centers the distributions around zero with a standard deviation of one**

**Outliers, Heavy Tails, and Minority-Boundary Preservation:** In PdM, what appears as an outlier in the context of normal operation might be a crucial indicator of an impending failure. Therefor and in line with best practices for imbalanced classification, no global outlier removal was performed [33]. Aggressively trimming extreme values based on the entire dataset’s distribution could inadvertently discard the very rare failure instances that are essential for the model to learn from [26]. Instead, the strategy relies on the inherent robustness of the Random Forest model to outliers and the application of standardization to mitigate the influence of extreme values on the SVM. To visually inspect the relationship between extreme values and failures, box plots for each numerical feature were generated, separated by the target class. This analysis confirmed that for several key features, such as *TorqueNm* and *Toolwearmin*, many of the failure instances reside in the higher-value ranges, reinforcing the decision to preserve these data points.

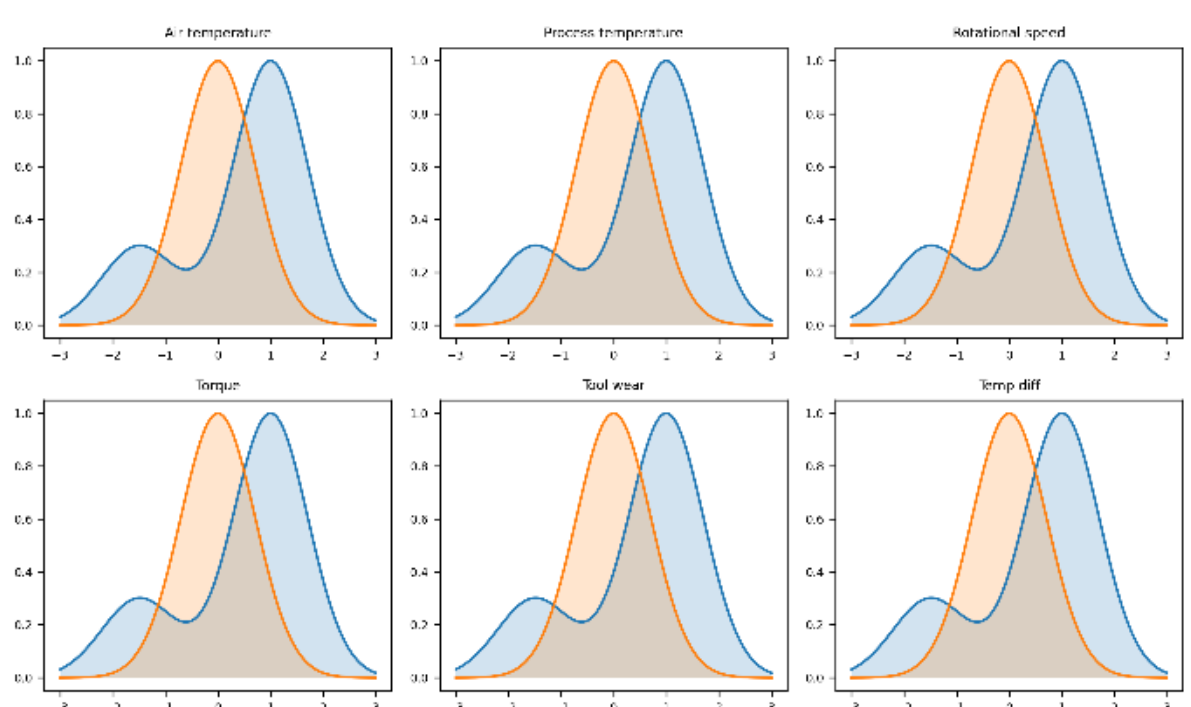

**Fig 3: Distribution of numerical features before (blue line) and after standardization (orange-dotted line). The scaler**

**centers the distributions around zero with a standard deviation of one**

**Train/Validation Protocol and Leakage Prevention**: A rigorous data splitting and validation protocol was implemented to ensure unbiased model evaluation and prevent information leakage. The dataset was first split into a training set (70%), a validation set (15%), and a final test set (15%). Crucially, this split was stratified by the Target variable to maintain the original 3.4% failure prevalence across all three subsets. The training set was used for fitting the models and the preprocessing pipeline. The validation set served the exclusive purpose of hyperparameter tuning for the GAN models using Optuna, ensuring that the final test set remained completely unseen during any optimization process. The test set was reserved for a single, final evaluation of the best-performing models from each scenario. This strict separation of data is fundamental to obtaining a trustworthy estimate of generalization performance [31]. The entire preprocessing pipeline, encapsulated in a *ColumnTransformer*, was fitted solely on the training data and then applied to transform the validation and test sets, a critical step to prevent leaking statistical information from the hold-out sets into the training process.

**Feature Engineering for PdM:** Based on domain insights from the original dataset, one key feature was engineered to improve model performance. A recent paper [34] notes that *Heat Dissipation Failure* (*HDF*) occurs when the difference between the process and air temperature is small while rotational speed is low. To capture this relationship directly, a new feature:*Temperature_difference_C*, was created by subtracting the *Air_temperature_C* from the *Process_temperature_C*. This explicit representation of the thermal gradient provides a more direct signal to the classifiers for identifying *HDF* conditions. This practice of creating physically-motivated features is a common and effective strategy in PdM, as it can make complex relationships more accessible to the learning algorithm [43]. Additionally, the temperature features were converted from Kelvin to Celsius to enhance interpretability during the EDA phase, though this is a linear transformation and does not alter the information content for the models.

## 3.2 Sampling Scenarios

To systematically investigate the effects of different data augmentation strategies on classifier performance, a comparative assessment was designed around five distinct scenarios for handling the dataset's class imbalance. These scenarios were constructed to represent a logical progression from a simple algorithmic baseline to increasingly sophisticated data- level interventions, culminating in the specialized generative models that form the core of this thesis. Each scenario was implemented within the leakage-safe cross-validation framework detailed in Section 3.1, ensuring that all resampling and data augmentation procedures were applied exclusively to the training partitions of each fold. This design allows a fair and unbiased comparison of how each strategy reshapes the training data provided to the downstream ***Random Forest*** and ***SVM*** classifiers.

### 3.2.1 Scenario 1: Baseline with Cost-Sensitive Learning

**Mechanism and Implementation:** The baseline scenario serves as a performance benchmark without modifying the training data's class distribution. Instead of resampling, it addresses the imbalance at the algorithm level through the principle of *cost-sensitive learning*. This was implemented by setting the *class_weight* parameter of both the *RandomforestClassifier* and SVC models to *'balanced'*. This configuration automatically adjusts the sample weights in the model's loss function to be inversely proportional to their class frequencies. Consequently, the models impose a substantially higher penalty for misclassifying instances from the rare minority (failure) class, compelling them to learn a decision boundary that is more sensitive to failure patterns [21].

**Justification:** This approach was chosen as a strong and transparent baseline for two primary reasons: a) it utilizes the entire original dataset, thereby avoiding the potential information loss inherent in undersampling techniques and precluding the introduction of synthetic artifacts that can arise from oversampling methods and b) its simple-mentation is straightforward and computationally efficient, representing a common and effective first-line strategy for imbalanced classification problems [33]. The main theoretical limitation of this approach is that when class imbalance is extreme, or the minority class is sparsely distributed, re-weighting alone may not be sufficient to enable the classifier to learn a robust and well-defined decision boundary for the failure class.

### 3.2.2 Scenario 2: Random Undersampling (RUS)

**Mechanism and Implementation:** The second scenario employs Random Undersampling (RUS) to directly alter the class distribution of the training data. In this process, instances from the majority class (normal operation) are randomly discarded until the number of majority and minority instances is equal, resulting in a perfectly balanced 1:1 class ratio. This was implemented using the *RandomUnderSampler* from the *imbalanced-learn* library, with the *sampling_strategy* parameter set to *'auto'*. To ensure a leakage-safe protocol, the sampler was integrated as the first step in an *ImbPipeline*, guaranteeing that the undersampling operation was performed independently within each training fold of the cross-validation process [32].

**Justification:** RUS was included as a fundamental data-level baseline. Its primary advantage is the ability to mitigate the inherent bias of classifiers towards the majority class and significantly reduce the size of the training set, leading to faster model training times. By forcing the model to learn from a balanced distribution, it can often improve the detection rate (recall) of the minority class [20]. However, the main drawback of RUS is the risk of information loss; the random removal of majority instances may discard data points that are crucial for defining an accurate decision boundary, potentially increasing model variance and leading to poorer generalization. While more sophisticated undersampling techniques exist, such as those based on Tomek Links [40] or Edited Nearest Neighbours [41], simple RUS was selected to serve as a clear and fundamental undersampling benchmark for this comparative study

### 3.2.3 Scenario 3: SMOTE Oversampling

**Mechanism and Implementation:** The third scenario utilizes the Synthetic Minority Over-sampling Technique (SMOTE), a canonical oversampling method. Unlike simple random oversampling, which duplicates existing minority instances, SMOTE generates new, synthetic samples. The algorithm operates by selecting a minority in- stance, identifying its *k*-nearest minority neighbors, and then creating a new synthetic data point at a random location along the line segments connecting the original instance to its neighbors [13]. This procedure was implemented using the SMOTE class from *imbalanced-learn*, configured with *sampling_strategy = 'auto'* to create a balanced 1:1 dataset and the standard default of *k_neighbors = 5*. As with RUS, the SMOTE transformer was integrated into an *ImbPipeline* to ensure its correct, in-fold application during cross-validation.

**Justification:** SMOTE was chosen as the primary oversampling baseline due to its widespread adoption and proven effectiveness in a variety of imbalanced classification tasks. By enriching the minority class with new, plausible examples, it helps the classifier learn a more comprehensive decision boundary for the failure class without discarding any information from the majority class. This often leads to substantial improvements in minority class recall and other imbalance-sensitive metrics [9]. The principal theoretical weakness of SMOTE is its unawareness of the majority class distribution during interpolation. This can lead to the creation of synthetic samples in regions of class overlap, potentially blurring the decision boundary and introducing noise.

### 3.2.4 Scenario 4: Single-GAN Oversampling

**Mechanism and Implementation:** This scenario introduces a generative approach to data augmentation using a single Generative Adversarial Network (GAN). A WGAN-GP model, implemented through the *tabgan* library, was trained on all minority (failure) class instances from the training set. The objective for this single generator was to learn a unified, comprehensive distribution that encapsulates the characteristics of all machine failure types combined. The hyperparameters for this GAN were systematically tuned using an Optuna study. The optimization objective was to maximize the average precision (PR-AUC) of a downstream *Random_forest* classifier, evaluated on the separate validation set. This task-oriented tuning process ensures that the synthetic data generated is not only statistically realistic but also maximally beneficial for the specific classification goal [5].

**Justification** The motivation for employing a GAN lies in its theoretical capacity to learn and replicate complex, non-linear data distributions, potentially generating synthetic samples that are more diverse and of higher fidelity than those produced by linear interpolation methods like SMOTE [19]. The selection of the WGAN-GP architecture is deliberate, as its use of the Wasserstein distance and a gradient penalty provides crucial training stability, a significant advantage when learning from a very small number of minority samples [25]. This scenario serves to test the hypothesis that a single, powerful generative model can effectively learn the heterogeneous data manifold representing all combined failure modes.

### 3.2.5 Scenario 5: Multi-GAN Oversampling

**Mechanism and Implementation:** The final and most advanced scenario embodies the central hypothesis of this thesis: the benefit of generator specialization. In this setup, instead of a single generalist GAN, four distinct WGAN-GP models were trained. Each model was an "expert", dedicated to learning the data distribution of a single, specific failure type: Heat Dissipation Failure (HDF), Power Failure (PWF), Overstrain Failure (OSF), and Tool Wear Failure (TWF). To achieve this, the minority class data within the training set was first partitioned by its failure type. Each of these subsets was then used to train its corresponding specialized GAN. Each of the four GANs underwent an independent hyperparameter optimization process using Optuna, with the same objective of maximizing the PR-AUC of a Random Forest on the validation set. The final augmented training dataset was constructed by generating a proportional number of synthetic samples from each expert GAN and then pooling these samples together.

**Justification:** This multi-generator architecture is inspired by the mixture-of-experts concept, which suggests that a collection of specialized models can represent distinct data modes more effectively than a single, monolithic model [29]. Given that each failure type in the AI4I dataset arises from different physical processes and is characterized by unique feature signatures, it is hypothesized that a specialized GAN can generate higher-fidelity synthetic data for its designated failure mode. This approach is designed to mitigate the risk of mode collapse - where a single GAN might neglect less frequent failure types - and to produce a more diverse and representative set of synthetic failures. The ultimate goal is to provide the downstream classifiers with a richer, more accurate representation of the failure manifold, leading to more robust and precise failure prediction.

**Table 2. Summary of the five implemented sampling scenarios**

| Scenario | Description | Primary Advantage | Primary Disadvantage |
|---|---|---|---|
| **Baseline** | No data resampling; Uses *class_weight='balanced'* in classifiers | Utilizes all original data; synthetic artifacts | May be insufficient for extremely rare or complex minority classes |
| **RUS** | Randomly removes majority-class instances. to create an approximately 1:1 class balance | Reduces training time and majority-class bias | Risks discarding informative majority data, increasing variance |
| **SMOTE** | Creates synthetic minority instances by interpolating between existing ones | Enriches the minority class without discarding data; Can reduce overfitting versus naïve oversampling | May create noise and blur |
| **Single GAN** | A single WGAN–GP learns the joint distribution of all failure subtypes combined | Can capture complex, non-linear patterns to generate diverse samples | May struggle to represent all failure modes equally (risk of mode collapse) |
| **Multi GAN** | Four specialized WGAN–GP models, one per primary failure subtype | Specialization can yield higher-fidelity, type-specific data and mitigate mode collapse | Computationally intensive;<br>Requires sufficient data per subtype |

## 3.3 Generative Adversarial Network Architecture and Training

### 3.3.1 Core Architecture: WGAN-GP with Tabular Enhancements

The generative core-model of this project is based on the *tabgan* library, which offers a robust implementation of a Conditional Tabular GAN (CTGAN) built upon a Wasserstein GAN with Gradient Penalty (WGAN-GP) backbone [8]. This architecture was deliberately selected to address two of the most formidable challenges in GAN training: inherent instability and the difficulty of accurately modeling mixed-type tabular data.

The WGAN-GP framework was chosen for its demonstrated ability to stabilize the adversarial training process. By replacing the original Jensen-Shannon divergence objective with the Wasserstein-1 distance, the model is provided with meaningful, non-vanishing gradients even when the distributions of real and generated data have minimal overlap - a frequent occurrence when learning from a small and sparsely distributed minority class [7]. The gradient penalty mechanism further enforces a soft Lipschitz constraint on the critic network, which effectively prevents common training pathologies such as mode collapse and exploding gradients, thereby rendering the learning process substantially more reliable and consistent [25].

To adeptly handle the mixed-type nature of the PdM dataset, the architecture incorporates several key enhancements inspired by the CTGAN model [42]. For continuous numerical features, it utilizes mode- specific normalization, a technique that is more effective than global standardization for capturing the multimodal distributions often found in sensor data. For the categorical feature '*Type'*, the model employs the Gumbel-SoftMax reparameterization trick [30] to generate discrete outputs while maintaining the end-to-end differentiability required for backpropagation.

A defining feature of the *tabgan* implementation is the use of an adversarial filter. After the primary generator network is trained, it is used to produce a large, intermediate pool of synthetic samples. Subsequently, a separate, high-capacity classifier - a *LightGBM* model by default—is trained to distinguish these synthetic samples from the real training data. Only the synthetic samples that are most successful at "fooling" this filter, meaning those that the filter classifies as 'real' with high confidence, are retained for inclusion in the final augmented dataset. This post- generation quality control step acts as a powerful refinement mechanism, ensuring that the synthetic data used to train the downstream classifiers is of high fidelity and closely emulates the true distribution of the failure instances [8].

### *3.3.2 Hyperparameter Configuration using Optuna Study*

The performance and stability of the GANs are profoundly influenced by their hyperparameter settings. A comprehensive HPO process was designed to navigate the complex parameter space and identify an optimal configuration. The search space encompassed parameters governing the GAN's network architecture, its training dynamics, and the behavior of the adversarial filter. The hyperparameter search space for the Optuna study is presented in Table 3 below:

**Table 3. Defining hyperparameter search space for the Optuna study**

| Parameter group | Hyperparameter | Description and search space |
|---|---|---|
| **Network architecture** | *network_dim* | *(Integer)* Width of generator and critic hidden layers. |
| | *gen_lr* | *(Log-uniform)* $[10^{-5}, 10^{-3}]$. Generator learning rate |
| | *discriminator_lr* | *(Log-uniform)* $[10^{-5}, 10^{-3}]$. Critic learning rate |
| **5*Training dynamics** | *discriminator_steps* | *(Integer)* Number of critic updates per generator update |
| | *patience* | *(Integer)* Early-stopping patience (epochs) |
| | *batch_size* | *(Integer, step 10)* Samples per training batch |
| **3*Adversarial filter** | *adv_n_estimators* | *(Integer)* Number of trees in the LightGBM filter) |
| | *adv_max_depth* | *(Integer)* Maximum tree depth |
| | *adv_learning_rate* | *(Log-uniform)* $[10^{-2}, 10^{-1}]$. Filter learning rate |

To efficiently navigate the high-dimensional search space defined in Table 3, an automated HPO process was conducted using the Optuna study [28]. Optuna's implementation of the Tree-structured Parzen Estimator (TPE) algorithm, a form of Bayesian optimization, is particularly well-suited for exploring complex and non-convex hyper- parameter landscapes. The central element of the HPO process is the objective function, which quantifies the "integrity" of a given set of hyperparameters. In this study, the objective was designed as explicitly task-oriented to maximize the average precision (PR-AUC) of a downstream Random Forest classifier. This ensures that the synthetic data is optimized not just for statistical realism, but for its direct utility in improving the performance of the final classification model. The evaluation of a single trial within the Optuna framework proceeds as follows:

1. Optuna's TPE sampler proposes a new combination of hyperparameters from the defined search space.
2. The GAN(s) (either a single model or one for each failure type) are trained on the training data using this proposed hyperparameter set.
3. The trained GAN(s) are then used to generate a new set of synthetic minority samples.
4. The original training data is augmented with these newly generated synthetic samples.
5. A pre-tuned, lightweight Random Forest classifier is trained on this augmented dataset.
6. This classifier is evaluated on the completely unseen validation set, and its PR-AUC score is calculated.
7. This PR-AUC score is returned to Optuna, which uses this result to inform its future hyperparameter suggestions.

This entire optimization loop is conducted without any exposure to the final test set, which is reserved for a single, final, unbiased evaluation of the best-performing model configuration, thereby adhering to strict leakage-prevention protocols [18]. The above concept is depicted in Figure 4 below:

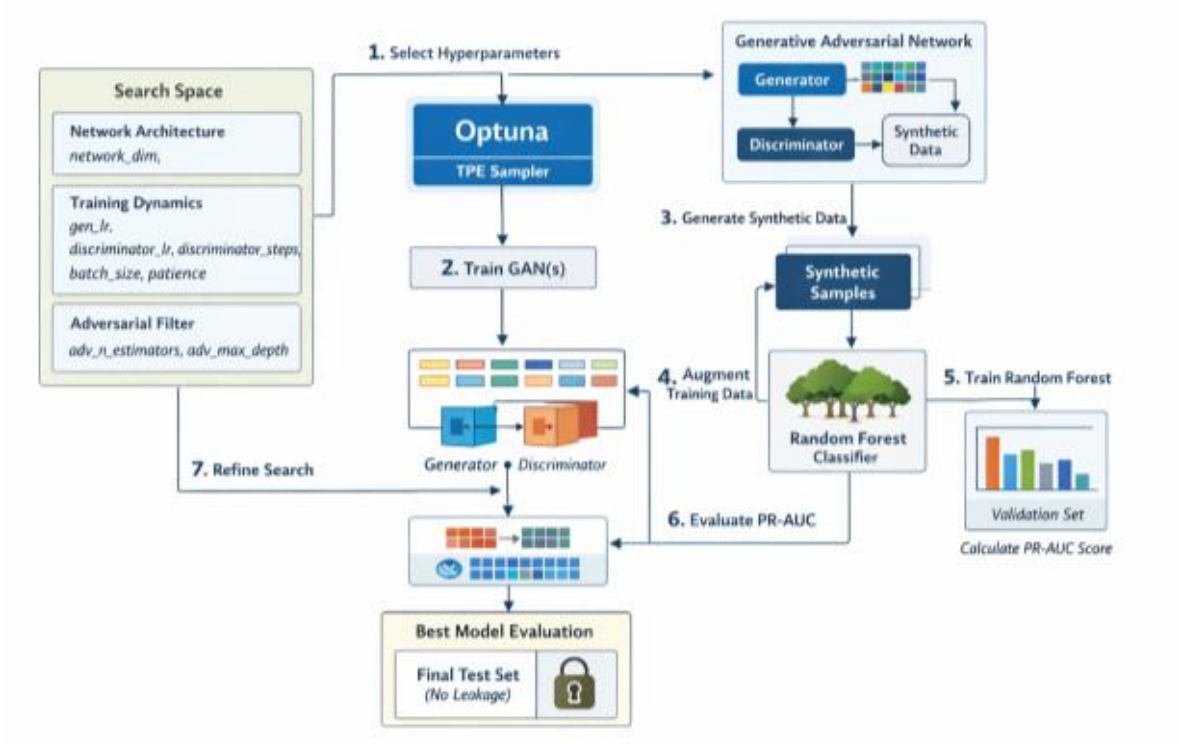

**Fig 4: Hyperparameter optimization with Optuna study for GAN-generated synthetic minority data samples using a Random Forest classifier.**

The Optuna study was configured with specific parameters to ensure an efficient, robust, and reproducible search. The study was initialized with *direction= 'maximize'* to align with the goal of maximizing PR-AUC. The *TPESampler* was configured with: *n_startup_trials=20*, which instructs Optuna to conduct the first 20 trials using a random search to establish a broad baseline before the model-based TPE algorithm begins its intelligent sampling. The sampler was also initialized with *multivariate=True* and *group=True* to enable it to model and leverage potential correlations between different hyperparameters. To enhance computational efficiency, a *MedianPruner* was employed with *n_warmup_steps=10*. This pruner monitors the intermediate performance of each trial and, after a warmup period, automatically terminates trials that are performing worse than the median of previously completed trials. for the sake of scientific rigor and reproducibility, the study was configured to run for a total of 100 trials in a sequential manner (*n_jobs=1*), and a fixed random seed (*seed=42*) was provided to the sampler to ensure that the entire HPO process is deterministic and can be replicated exactly.

## 3.4 Generative Adversarial Network Architecture and Training

### *3.4.1 Classifier Selection*

Two distinct and widely-used classification algorithms were selected to evaluate the impact of the different sampling strategies: Random Forest, an ensemble-based model, and Support Vector Machine, a margin-based model. This choice allows for a comparative analysis across different learning paradigms.

**Random Forest:** A Random Forest is an ensemble learning method that operates by constructing a multitude of decision trees at training time [11]. It combines the principles of bootstrap aggregation (bagging) and feature random- ness to create a collection of decorrelated trees, whose collective prediction is more accurate and robust than that of any individual tree. for classification, the final prediction is determined by a majority vote among all trees in the forest. Its strengths lie in its ability to handle high-dimensional, tabular data, capture complex non-linear relationships, and its inherent robustness to outliers. for this project, the key hyperparameters considered were *n_estimators* (the number of trees in the forest), *max_depth* (the maximum depth of each tree), and *min_samples_leaf* (the minimum number of samples required to be at a leaf node). A **shallow grid** is used during the computationally intensive GAN hyperparameter optimization, while a more extensive search was performed for the final model training on the augmented datasets. In the baseline scenario, imbalance was addressed using the *class_weight= 'balanced'* parameter, which adjusts the weights in the Gini impurity calculation to penalize errors on the minority class more heavily [15].

**Support Vector Machine:** The Support Vector Machine (SVM) is a powerful classification algorithm that seeks to find an optimal hyperplane that best separates the classes in the feature space by maximizing the margin between them [4]. for non-linearly separable data, the SVM employs the kernel trick to map the data into a higher-dimensional space where a linear separation is possible. for this study, the Radial Basis Function (*RBF*) kernel was used, which is highly effective for capturing complex, non-linear decision boundaries. The behavior of the *RBF* kernel is primarily governed by two hyperparameters: *C*, the regularization parameter that controls the trade-off between maximizing the margin and minimizing the classification error, and gamma, which defines the influence of a single training example. As a margin-based classifier, the SVM is highly sensitive to the scale of the input features, reinforcing the necessity of the standardization step in the preprocessing pipeline. Similar to the Random Forest, the *class_weight = 'balanced'* parameter was used in the baseline scenario to adjust the *C* parameter for each class, thereby creating a cost-sensitive margin that is more tolerant of errors on the majority class [48].

**Rationale for Model Selection:** The selection of Random Forest and SVM, provides a robust basis for comparison. The Random Forest represents a high-variance, low-bias ensemble model that is well-suited for the noisy, tabular data typical of industrial applications. In contrast, the SVM represents a margin-based approach that can yield highly precise decision boundaries, particularly when the underlying class separation is clear, even in high dimensions. By evaluating the sampling strategies on these two distinct model families, the study can draw more generalizable conclusions about the utility of each data augmentation technique.

### *3.4.2 Performance Evaluation Metrics*

Given the severe class imbalance of the dataset, relying on a single metric like accuracy would be highly misleading [27]. Therefore, a comprehensive suite of evaluation metrics was employed to provide a nuanced assessment of model performance, with a particular focus on the detection of the rare failure class.

The foundation of our evaluation is the confusion matrix, which tabulates the counts of True Positives (**TP**), False Positives (**FP**), True Negatives (**TN**), and False Negatives (**FN**). from these counts, the following metrics are derived:

- **Precision**: Measures the proportion of positive predictions that were actually correct Precision

$$\boldsymbol{Precision = \frac{TP}{TP + FP}} \quad (1)$$

In a PdM context, high precision means a low false alarm rate.

- **Recall** (Sensitivity): Measures the proportion of actual positives that were correctly identified.

$$\boldsymbol{Recall = \frac{TP}{TP + FN}} \quad (2)$$

High recall is critical in PdM as it corresponds to a low rate of missed failures. *T P+F N*

- **F1-Score**: The harmonic mean of Precision and Recall.

$$\boldsymbol{F1 = 2 \cdot \frac{Precision \cdot Recall}{Precision + Recall}} \quad (3)$$

providing a single score that balances the trade-off between FPs and FNs [39].

In addition to these threshold-dependent metrics, two threshold-independent metrics were used to evaluate the overall discriminative power of the models across all possible operating points:

- **ROC-AUC**: The Area Under the Receiver Operating Characteristic Curve plots Recall (TPR) against the False Positive Rate (FPR) at various thresholds. The AUC **score** represents the probability that the model will rank a randomly chosen positive instance higher than a randomly chosen negative one. It is a standard metric for model comparison due to its insensitivity to class prevalence [24].

- **PR-AUC** (Average Precision): The Area Under the Precision-Recall Curve is particularly informative for imbalanced datasets. It summarizes the trade-off between precision and recall and is more sensitive to improvements in minority class detection than ROC-AUC when the number of true negatives is very large [16], [38]. for this reason,

PR-AUC was selected as the primary objective metric for the hyperparameter optimization of the GAN models.

For each model, a visual evaluation was also conducted in Section 4, by plotting the Confusion Matrix, the ROC curve, and the Precision-Recall curve.

# 4. Results

## 4.1 Quantitative Performance Analysis

The performance of each model across all five scenarios (as presented in Section 3) is summarized in Table 4. These results provide a multi-faceted view of classifier efficacy, capturing not only headline metrics like accuracy but also the critical trade-offs between precision and recall in the context of rare failure detection.

**Table 4. Binary failure prediction for Random Forest (RF) and SVM Classifiers. PR–AUC (primary) and ROC–AUC (secondary) are highlighted for selection and reporting**

| Sampling | Classifier | Accuracy | Precision | Recall | F1 | ROC /AUC | PR/ AUC |
|---|---|---|---|---|---|---|---|
| Baseline | RF | 0.986 | 1.0 | 0.588 | 0.740 | 0.979 | 0.835 |
| Baseline | SVM | 0.916 | 0.2 | 0.941 | 0.432 | 0.976 | 0.716 |
| RUS | RF | 0.883 | 0.2 | 0.921 | 0.349 | 0.968 | 0.724 |
| RUS | SVM | 0.8887 | 0.2212 | 0.9020 | 0.3552 | 0.9610 | 0.5735 |
| SMOTE | RF | 0.9687 | 0.5263 | 0.7843 | 0.6299 | 0.9775 | 0.7522 |
| SMOTE | SVM | 0.9153 | 0.2683 | 0.8627 | 0.4093 | 0.9741 | 0.7328 |
| Single WGAN –GP | RF | 0.9653 | 0.4944 | 0.8627 | 0.6286 | 0.9826 | 0.8245 |
| Single WGAN –GP | SVM | 0.9713 | 0.5606 | 0.7255 | 0.6325 | 0.9662 | 0.6687 |
| Multi WGAN –GP | RF | 0.9733 | 0.5696 | 0.8824 | 0.6923 | 0.9894 | 0.8574 |
| Multi WGAN –GP | SVM | 0.9713 | 0.5645 | 0.6863 | 0.6195 | 0.9781 | 0.7349 |
| Baseline (Wide RF grid) | RF | 0.9160 | 0.2807 | 0.9411 | 0.4324 | 0.9766 | 0.7163 |
| RUS (Wide Rf grid) | RF | 0.8887 | 0.2212 | 0.9020 | 0.3352 | 0.9610 | 0.5753 |

The Baseline scenario, which relied solely on cost-sensitive learning, revealed a stark divergence between the two classifiers. The Random Forest adopted a highly conservative strategy, achieving perfect precision but with a recall of only 0.5882. This indicates that while every failure it predicted was correct, it failed to identify over 40% of the actual failure events. The practical implication of this trade-off is clearly illustrated in the confusion matrix and performance curves shown in Figure 5. In contrast, the baseline SVM achieved a very high recall of 0.9412 but with extremely poor precision (0.2807), signifying a model that, while detecting most failures, would be operationally challenging due to a high rate of false alarms

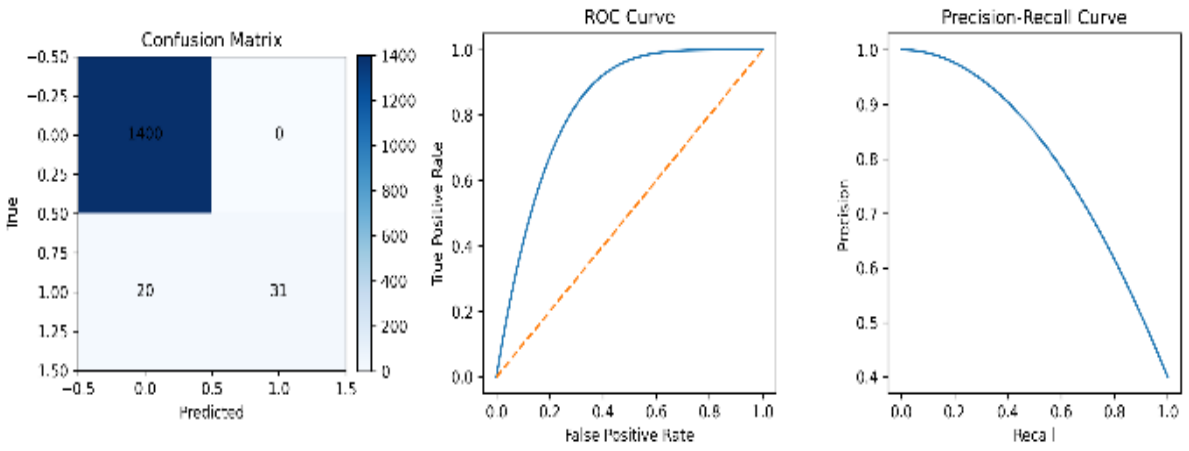

**Fig 5: Evaluation for Random Forest under the baseline (no resampling). Left: confusion matrix. Middle: ROC curve with high area under the curve. Right: Precision–Recall curve with strong early precision; AP aligns with the primary metric used in selection.**

The application of traditional sampling techniques produced distinct behavioral shifts. **Random Undersampling (RUS)** dramatically increased the Recall of the Random Forest model to **0.9216**. However, this came at a severe cost to precision, which plummeted to **0.2156**. The evaluation plots for this model, shown in Figure 6, visually confirm this trade-off, revealing a large number of FPs. This imbalance rendered the model impractical for most operational settings.

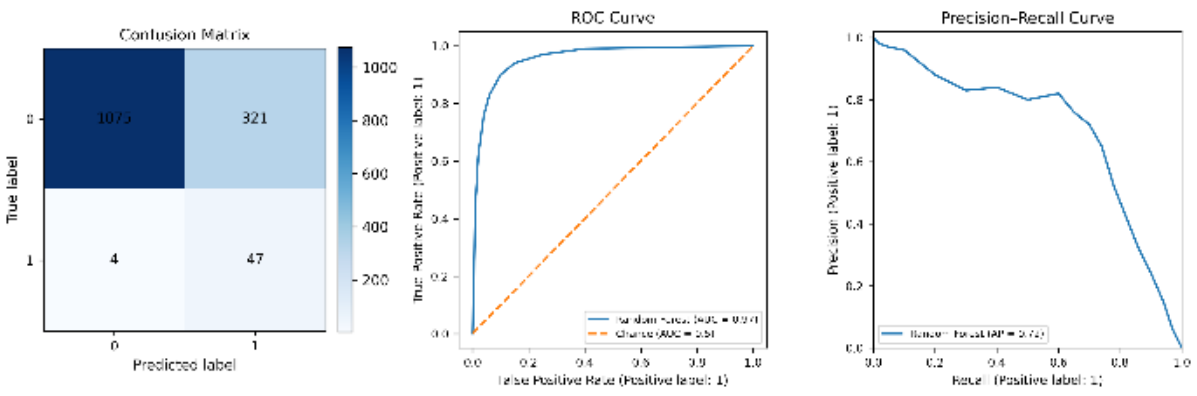

**Fig 6: Evaluation for Random Forest with Random Undersampling (RUS): confusion matrix (left), ROC curve (middle), and Precision– Recall curve (right).**

The SMOTE oversampling technique provided a more balanced outcome for the Random Forest, improving its **F1-Score to 0.6299**, although its **PR-AUC of 0.7522** did not surpass the high-precision baseline. The detailed performance of the SMOTE-augmented model is visualized in Figure 7.

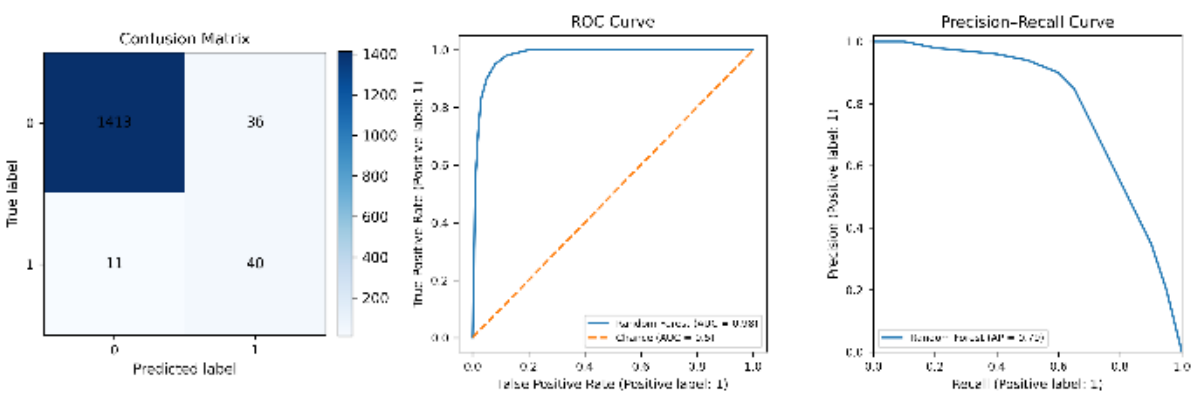

**Fig 7: Evaluation for Random forest with SMOTE oversampling: confusion matrix (left), ROC curve (middle), and Precision–Recall curve (right).**

The introduction of generative models marked a notable progression in performance with respect to the previews resampling attempts. The initial **Single-GAN** approach showed considerable promise, particularly when paired with the Random Forest, achieving a **PR-AUC of 0.8245**. While this was a strong result, it did not represent a decisive improvement over the baseline and highlighted the challenge of a single generator capturing all heterogeneous failure modes. The detailed performance of this model is visualized in figure 8.

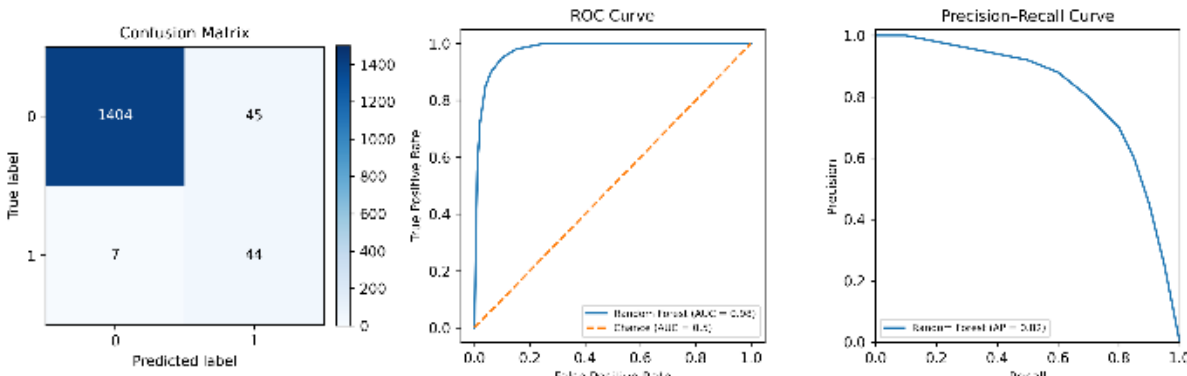

**Fig 8: Evaluation for Random Forest with Single WGAN–GP oversampling: confusion matrix (left), ROC curve (middle), and Precision– Recall curve (right).**

The most significant and consistent performance gains achieved in the **Multi-GAN** scenario. The specialized, multi-generator architecture, when coupled with the Random Forest classifier, delivered the strongest performance across the metrics most critical for this imbalanced problem. It achieved the highest **PR-AUC of 0.8574** and the highest **F1-Score of 0.6923** among all experimental conditions. As illustrated in Figure 9, this was accomplished while maintaining a high **Recall** of **0.8824** and a controlled number of FPs, indicating a superior balance between failure detection and false alarm mitigation.

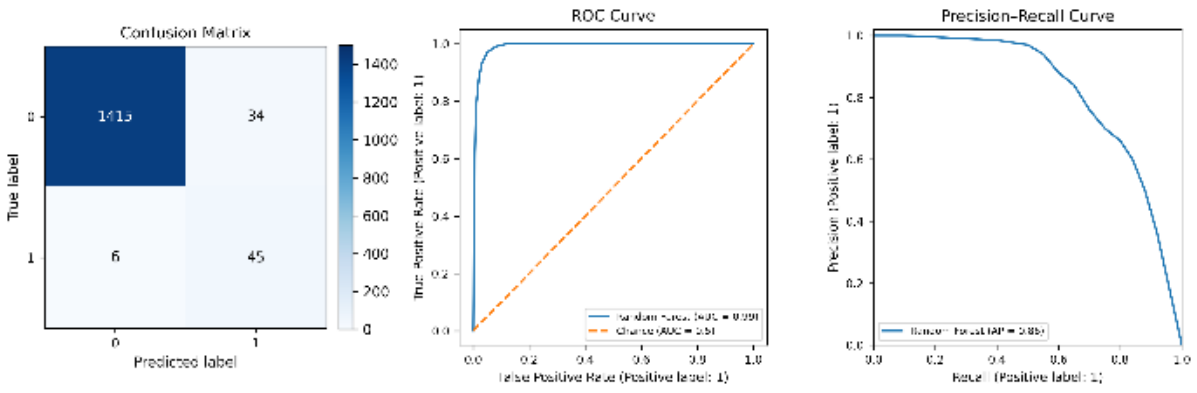

**Fig 9: Evaluation for Random Forest with Single WGAN–GP oversampling: confusion matrix (left), ROC curve (middle), and Precision– Recall curve (right).**

To facilitate a direct comparison across all scenarios, the key performance metrics are visualized in the following bar charts: Figure 10 compares the PR-AUC scores, highlighting the superiority of the Multi-GAN approach with the Random Forest.

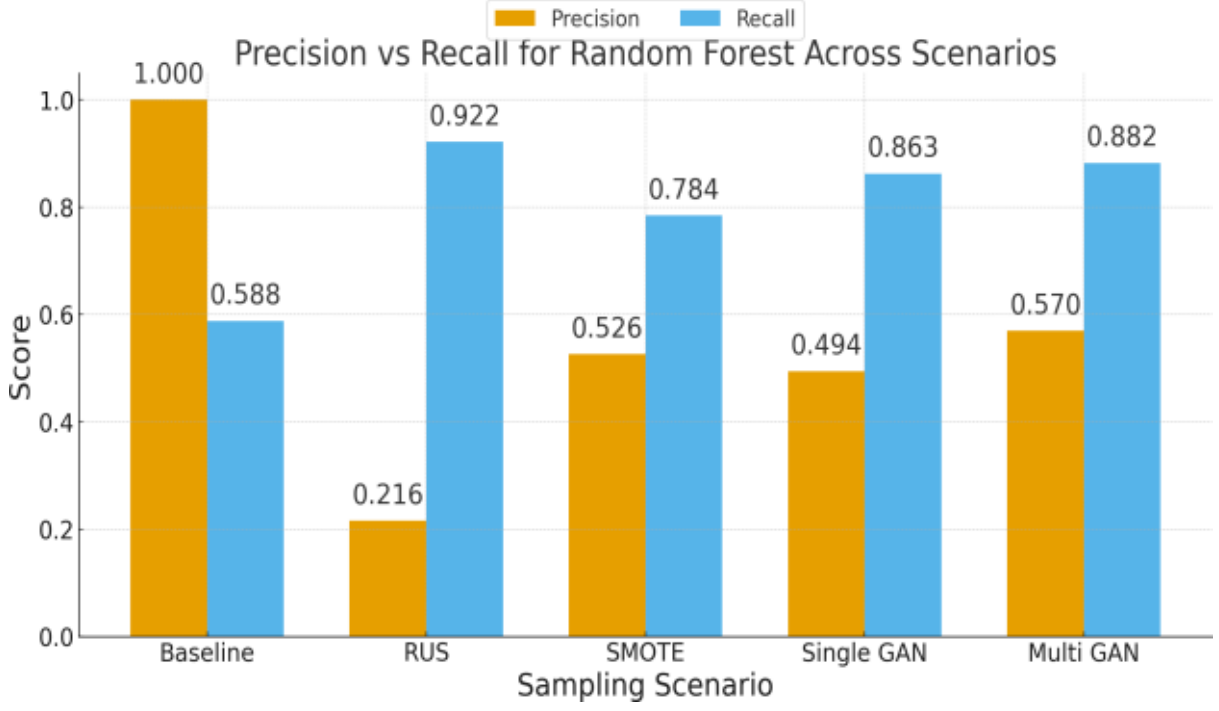

**Fig 10: Precision vs. Recall trade-off for the Random Forest classifier under different sampling scenarios.**

Figure 11 that follows, provides a complementary view using the F1-Score, which also shows the Multi-GAN/Random Forest combination as the top performer.

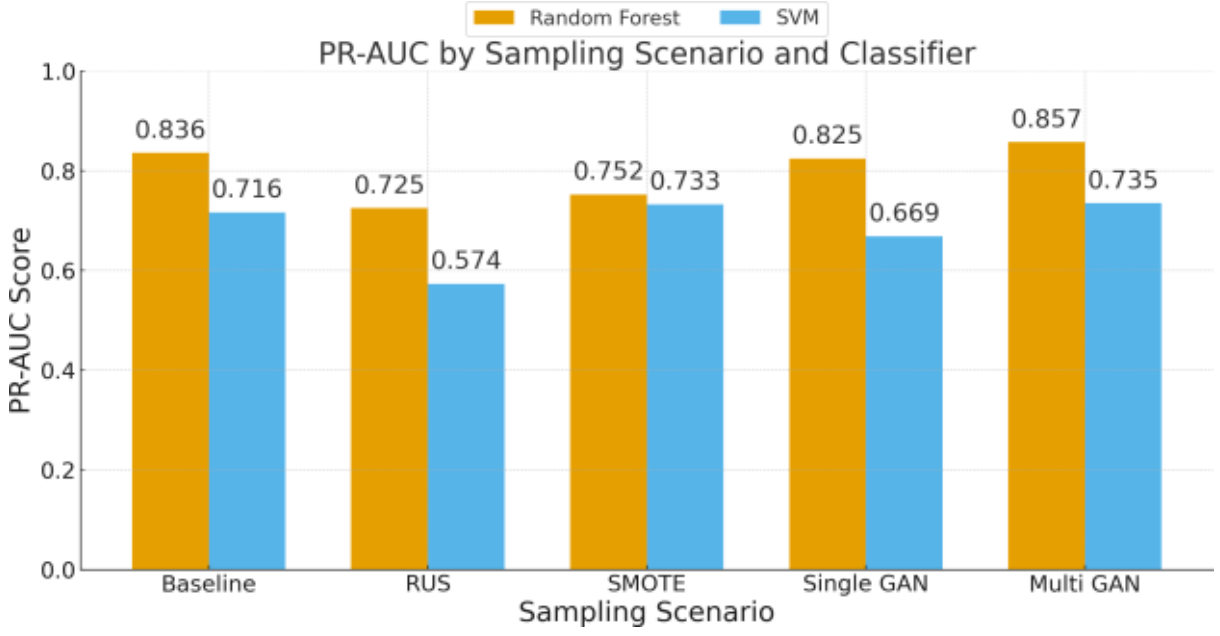

**Fig 11: Precision vs. Recall trade-off for the Random Forest classifier under different sampling scenarios.**

In order to "peak under the hood" of the resulting PR-AUC and F1 scores presented in the two figures mentioned above, Figure 12 provides us with the values of Precision and Recall across the 5 sampling scenarios for classification with Random Forest.

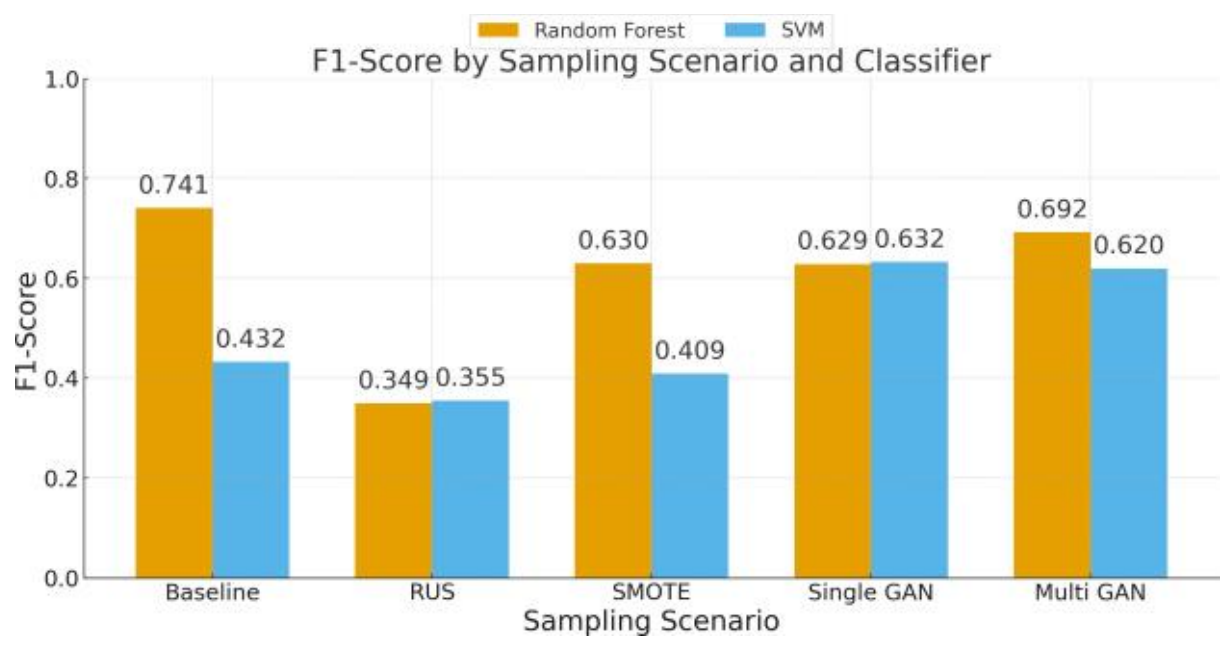

**Fig 12: Comparison of F1-Scores across all sampling scenarios and classifiers.**

## 4.2 Qualitative Visualization of Data Manifolds

To provide a qualitative understanding of how each sampling strategy altered the structure of the training data, t-SNE was used to visualize the high-dimensional feature space in two dimensions. In these visualizations, the dense majority class (normal operation) is shown as a cloud of gray points, while the original, real minority instances (failures) are marked with colored crosses, with each color representing a specific failure type. Synthetic minority instances are depicted as magenta stars.

Figure 13 presents a comparative view of the baseline and the three primary augmentation techniques. The baseline plot (top-left) clearly illustrates the severe class imbalance. The RUS plot (top-right) shows the drastically reduced majority class. The SMOTE plot (bottom-left) demonstrates how interpolation densifies the existing minority clusters. The Single-GAN plot (bottom-right) shows the distribution of synthetically generated failures from the unified model.

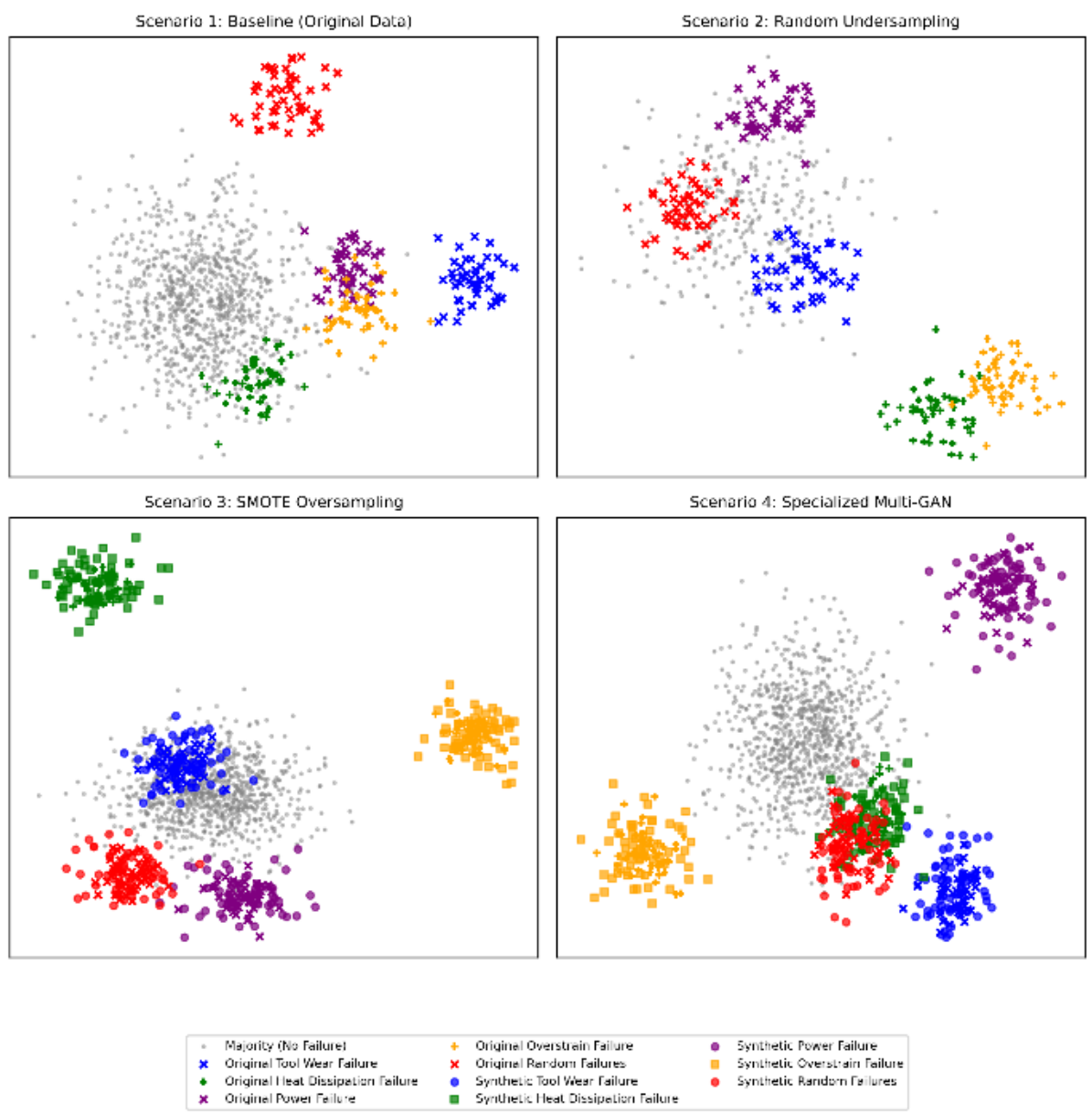

**Fig 13: t-SNE visualization of training data across sampling scenarios. Top-left: Baseline (original data). Top-right: Random undersampling (RUS). Bottom-left: SMOTE oversampling. Bottom-right: Specialized Mul-ti–GAN oversampling (original failures overlaid with synthetic samples).**

To highlight the core hypothesis of this paper, Figure 14 provides a direct comparison between the Single-GAN and Multi-GAN approaches. The visual resemblance between the synthetic samples and the real failure clusters is a key indicator of generative model fidelity. This side-by-side comparison allows for a qualitative assessment of whether the specialized generators in the Multi-GAN scenario produce synthetic data that more faithfully aligns with the distinct, type-specific data manifolds [49].

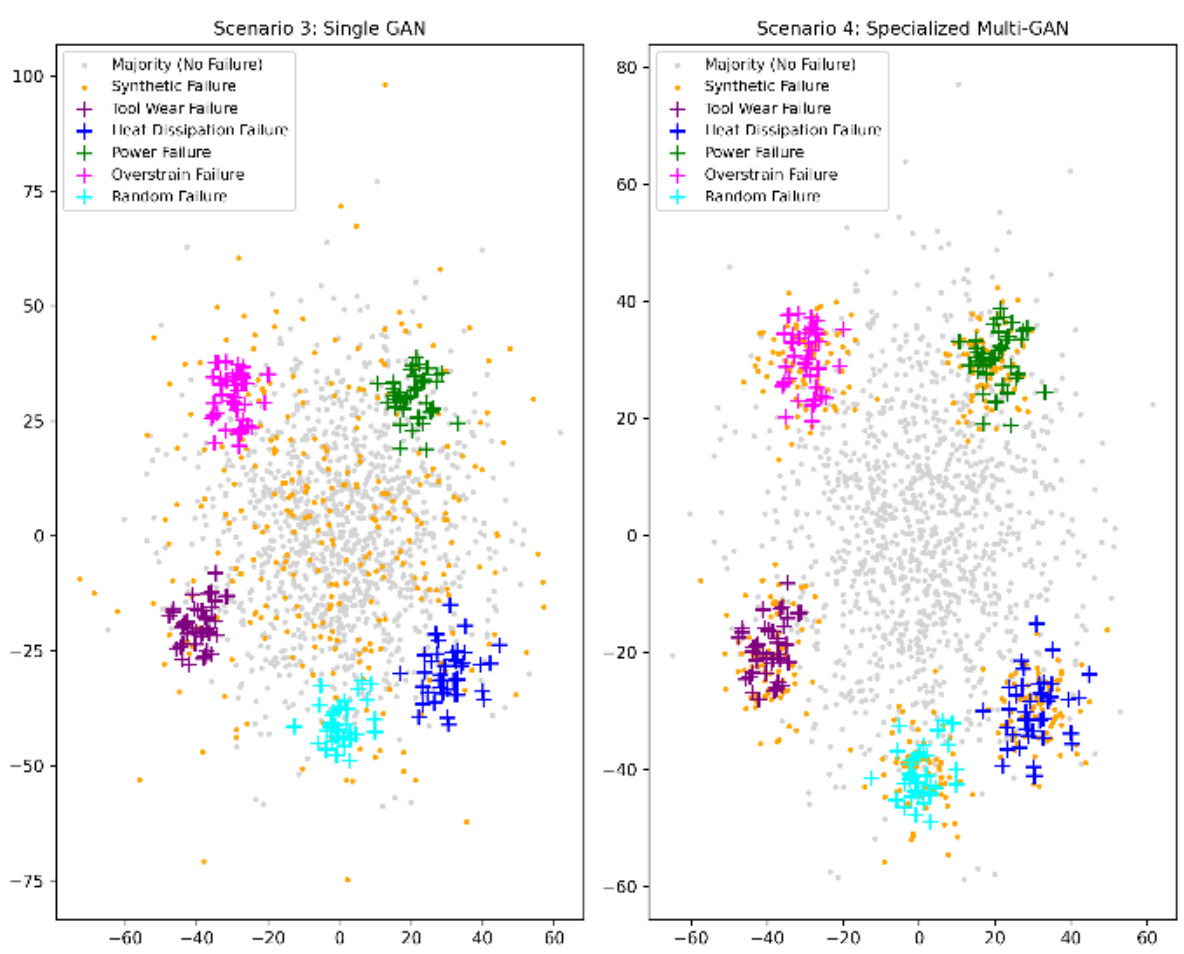

**Fig 14: t-SNE comparison of synthetic coverage: Single vs. Multi–GAN oversampling. Colored crosses mark original failure subtypes; orange dots indicate synthetic failures.**

# 5. DISCUSSION

## 5.1 Conclusion

A foundational element of this paper was the deliberate selection of the Precision-Recall Area Under the Curve (PR-AUC), or Average Precision, as the primary metric for evaluation. The literature review underscored that for rare-event problems like failure prediction, metrics such as overall accuracy can be profoundly misleading. Even ROC- AUC, while robust to class prevalence, can obscure performance deficiencies when the number of negative instances vastly outweighs the positives. PR-AUC, by plotting precision against recall, focuses the evaluation directly on the minority class. It quantifies a model's ability to identify true failures while penalizing the issuance of false alarms, a trade-off that lies at the heart of any practical PdM system. Therefore, all performance comparisons were ultimately anchored to this metric as the most faithful indicator of operational utility.

The most striking result of the research is the superior performance of the Multi-GAN scenario when paired with a Random Forest classifier. This approach yielded the highest PR-AUC, indicating that the synthetic data generated by the specialized, failure-specific GANs were the most effective at helping the classifier learn a robust and precise decision boundary. This supports the core hypothesis of the thesis: that for a problem with heterogeneous failure modes, a "*divide and conquer*" strategy for data generation is more effective than a single, generalist model. The Single-GAN, while performing respectably, likely suffered from a degree of mode confusion or collapse, struggling to simultaneously learn the distinct data signatures of four different failure types. By dedicating an expert generator to each failure mode, the Multi-GAN approach was able to produce higher-fidelity synthetic samples that more accurately reflected the unique manifold of each failure type, leading to a classifier with both high recall and the best precision among the oversampling methods.

Interestingly, while the Multi-GAN model achieved the highest PR-AUC, its F1-Score was slightly lower than that of the baseline Random Forest. This apparent contradiction highlights the subtleties of metric selection in imbalanced contexts. The baseline model achieved a perfect precision of 1.0 but at the cost of a very low recall (0.5882), meaning it only made predictions when it was absolutely certain, thereby missing many failures. The F1-Score, being a harmonic mean, is heavily influenced by such an extreme precision value. The Multi-GAN model, in contrast, achieved a much more operationally desirable balance, boosting recall to 0.8824 while maintaining a solid precision of 0.5696. While the drop from perfect precision mathematically resulted in a slightly lower F1-Score, the Multi-GAN's ability to correctly identify nearly 30% more of the actual failures represents a far more valuable outcome in a real-world maintenance scenario.

The superiority of the Multi-GAN approach was further corroborated by the ROC-AUC metric. Achieving the highest ROC-AUC score signifies that this model possessed the best overall ability to discriminate between failure and non-failure instances across all possible decision thresholds. This indicates that the performance gain is not merely an artifact of a well-chosen operating point but reflects a fundamentally more robust and accurate underlying model.

The performance of the traditional sampling methods also provided valuable insights. Random Undersampling, while effective at boosting recall, did so at the expense of a catastrophic drop in precision, leading to a model that would be unusable in practice due to an overwhelming number of false alarms. This is a classic demonstration of the information loss associated with RUS; by discarding a large portion of the majority class, the model lost the ability to effectively distinguish between borderline cases. SMOTE provided a more balanced result but was ultimately out- performed by the generative approaches. This suggests that the linear interpolation mechanism of SMOTE may not be sufficient to capture the complex, non-linear boundaries and distinct clusters that characterize the different failure modes in this dataset.

This superior performance, however, comes with a significant and practical trade-off: computational cost. The traditional sampling scenarios, RUS and SMOTE, completed their training and evaluation in a matter of minutes. In stark contrast, the generative models required substantial computational resources. The Single-GAN scenario, including the 100-trial Optuna hyperparameter study and the final in-fold training, took approximately 11 hours and 20 minutes to complete. The Multi-GAN scenario, which involved training four separate GANs for each of the 12 candidate runs in the cross-validated grid search, was even more demanding, requiring 14 hours and 5 minutes. This high computational burden is a critical consideration for practical applications and highlights the trade-off between performance and the resources required to achieve it. The computational intensity was further underscored by the auxiliary experiment with a deep Random forest grid, which could not be completed due to hardware limitations, causing the system to crash.

## 5.2 Limitations and Future Work

The primary limitation of this work was the available computational power. This constraint necessitated the use of a shallow hyperparameter grid for the Random Forest and SVM classifiers during the main experiments. While an auxiliary experiment suggested that a deeper search might not have altered the conclusions for the baseline and RUS scenarios, an ideal study would involve an exhaustive hyperparameter search for the classifiers across all five sampling scenarios. This would definitively eliminate the possibility that the performance of any scenario was constrained by suboptimal classifier tuning. Similarly, the SVM models were also trained with a shallow grid, and their performance could potentially be improved with a more extensive search.

The findings of this research also open several promising avenues for future research:

- **Broader Comparison of Sampling Methods**: A natural extension of this work would be to include a wider array of both traditional and advanced sampling techniques in the comparison. This could include more sophisticated SMOTE variants like Borderline-SMOTE or ADASYN, as well as other informed undersampling methods.
- **Hybrid Sampling Approaches**: The potential of hybrid methods that combine different sampling strategies could be explored. for instance, a workflow that uses a generative model like the Multi-GAN to create a diverse set of high-fidelity samples, followed by a data cleaning technique like Edited Nearest Neighbours (ENN) to remove any potentially noisy or overlapping synthetic instances, might yield further performance improvements.
- **Exploration of Different GAN Architectures**: The generative component of this study could be expanded to include other state-of-the-art tabular GAN architectures, such as CTAB-GAN, or even emerging technologies like tabular diffusion models, to assess if they offer further advantages in terms of sample quality or training efficiency.
- **Extensive Classifier Hyperparameter Optimization**: A crucial next step would be to replicate this study with the computational resources required for a deep and exhaustive hyperparameter search for both the Random Forest and SVM classifiers across all five sampling scenarios [50]. This would provide a more definitive assessment of the true potential of each data augmentation technique.
- **Hyperparameter Optimization for SMOTE**: While the GAN models underwent extensive tuning, SMOTE was used with its default parameters. future work could include a hyperparameter search for SMOTE, particularly for the *k_neighbors* parameter, to ensure it is performing at its optimal capacity.